# Learning Graph-Structured Sum-Product Networks for Probabilistic Semantic Maps

**Kaiyu Zheng, Andrzej Pronobis and Rajesh P. N. Rao**\*

### Abstract

We introduce Graph-Structured Sum-Product Networks (GraphSPNs), a probabilistic approach to structured prediction for problems where dependencies between latent variables are expressed in terms of arbitrary, dynamic graphs. While many approaches to structured prediction place strict constraints on the interactions between inferred variables, many real-world problems can be only characterized using complex graph structures of varying size, often contaminated with noise when obtained from real data. Here, we focus on one such problem in the domain of robotics. We demonstrate how GraphSPNs can be used to bolster inference about semantic, conceptual place descriptions using noisy topological relations discovered by a robot exploring large-scale office spaces. Through experiments, we show that GraphSPNs consistently outperform the traditional approach based on undirected graphical models, successfully disambiguating information in global semantic maps built from uncertain, noisy local evidence. We further exploit the probabilistic nature of the model to infer marginal distributions over semantic descriptions of as yet unexplored places and detect spatial environment configurations that are novel and incongruent with the known evidence.

## Introduction

Graph-structured data appear in a wide range of domains, from social network analysis (Mislove et al. 2010), to computer vision (Johnson et al. 2015) and robotics. Often, the global structure of such data varies, yet dependencies captured by elements of the structure persist and can serve as a powerful source of information for various inference tasks. In robotics, this phenomenon is common. While exploring their environments, robots build a growing body of knowledge captured at different spatial locations, scales, and at different levels of abstraction. However, importantly, they also perceive relations between such information, which become invaluable in real-world settings where inference is performed with noisy data and under partial observability.

\*The authors are with Paul G. Allen School of Computer Science & Engineering, University of Washington, Seattle, WA, USA. A. Pronobis is also with Robotics, Perception and Learning Lab, KTH, Stockholm, Sweden. {kaiyuzh,pronobis,rao}@cs.washington.edu. This work was supported by Office of Naval Research (ONR) grant no. N00014-13-1-0817, the Keck foundation, and Swedish Research Council (VR) project 2012-4907 SKAEENet.

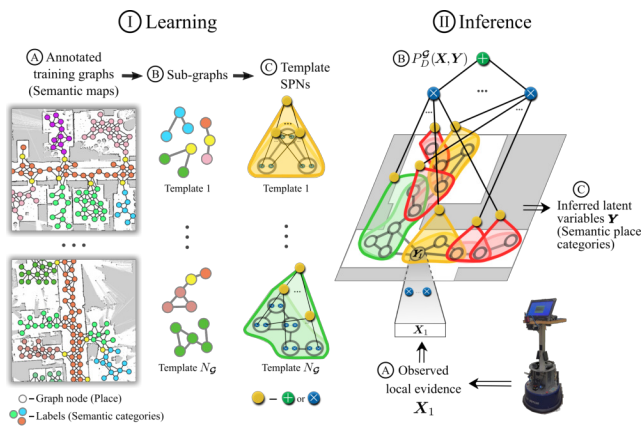

Figure 1: Learning and inference with GraphSPNs for probabilistic semantic maps with topological spatial relations.

Semantic maps are an established framework for modeling relational spatial knowledge in robotics (Friedman, Pasula, and Fox 2007; Zender et al. 2008; Pronobis and Jensfelt 2012). They often form graph-structured environment representations by utilizing topological graphs consisting of place nodes connected with local spatial relations (see Fig. 1 for an example). This enables anchoring of high-level semantic information about places and forms an accessible belief state representation for a planning algorithm (Hanheide et al. 2017). In order to integrate the collected spatial knowledge, resolve ambiguities, and make predictions about unobserved places, semantic mapping frameworks often employ structured prediction algorithms (Friedman, Pasula, and Fox 2007; Mozos et al. 2007; Pronobis and Jensfelt 2012). Unfortunately, the relations discovered by a robot exploring a real-world environment tend to be complex and noisy, resulting in difficult inference problems. At the same time, topological graphs are dynamic structures, growing as the robot explores its environment, and containing a different number of nodes and relations for every environment. Yet, many approaches to structured prediction place strict constraints on the interactions between inferred variables to achieve tractability (Bach and Jordan 2002) and require that the number of output latent variables be constant and related through a similar global structure (Belanger and McCallum 2016). This makes them either unsuitable or impractical in robotics settings without compromising on the structure complexity (Mozos et al. 2007), introducing prior structural knowledge (Friedman, Pasula, and Fox 2007), or making hard commitments about values of semantic attributes (Pronobis and Jensfelt 2012). These problems are not unique to robotics and often present themselves in other domains, such as computer vision (Schwing and Urtasun 2015).

In this paper, we present Graph-Structured Sum-Product Networks (GraphSPNs), a general probabilistic framework for modeling graph-structured data with complex, noisy dependencies between a varying number of latent variables. Our framework builds on Sum-Product Networks (SPNs) (Poon and Domingos 2011), a probabilistic deep architecture with solid theoretical foundations (Peharz et al. 2017). SPNs can learn probabilistic models capable of representing context-specific independence directly from high-dimensional data and perform fast, tractable inference on high-treewidth models. As illustrated in Fig. 1, GraphSPNs learn template SPN models representing distributions over attributes of sub-graphs of arbitrary complexity. Then, to perform inference for a specific, potentially expanding graph, they assemble a mixture model over multiple decompositions of the graph into sub-graphs, with each sub-graph modeled by an instantiation of an appropriate template.

We apply GraphSPNs to the problem of modeling large-scale, global semantic maps with noisy topological spatial relations built by robots exploring multiple office environments. We make no assumptions about the structure of the topological map that would simplify the inference over semantic attributes. Our approach is capable of disambiguating uncertain and noisy local information about semantic attributes of places as well as inferring distributions over semantic attributes for yet unexplored places, for which local evidence is not available. Furthermore, we illustrate the benefits of the probabilistic representation by relying on likelihood of global semantic maps to detect novel and incongruent spatial configurations. We compare the performance of our model with the traditional approach based on Probabilistic Graphical Models assembled according to the structure of the topological graph. We show that GraphSPNs significantly outperforms Markov Random Fields built from pairwise and higher-order potentials relying on the same uncertain evidence. Finally, we contribute a dataset[1] of topological graphs associated with evidence about semantic place attributes with different levels of uncertainty and varying amount of noise. We hope that the dataset will become a useful benchmark for evaluating approaches to modeling graph-structured data.

## Related Work

Probabilistic graphical models (PGMs) (Koller and Friedman 2009) provide a flexible framework for structured prediction. PGMs express the conditional dependence between random variables in terms of a graph. This results in a factorization of the joint distribution into a normalized products of factors, with factors defined over subsets of variables. The factors can be considered templates from which a distribution over an arbitrary number of variables can be assembled. As a result, PGMs were used in the past for modeling graph-structured data (Friedman, Pasula, and Fox 2007; Pronobis and Jensfelt 2012). Unfortunately, inference in PGMs is generally intractable, with the exception of low treewidth models (Bach and Jordan 2002). Even models with pairwise potentials require approximate inference techniques with no guarantee of convergence, such as Loopy Belief Propagation (BP) (Murphy, Weiss, and Jordan 1999), when the graph structure involves loops (as in this work). Higher-order models can pose a challenge even to the approximate methods. Additionally, many distributions which can be represented compactly, cannot be represented in terms of the factorization performed by PGMs (Poon and Domingos 2011).

An increasing number of structure prediction approaches utilize deep architectures. Several approaches build on graphical models for representing dependencies between output variables, while relying on deep models to realize the factors (Schwing and Urtasun 2015; Chen et al. 2015). Other types of models focus specifically on sequential data by combining local convolutions with classification (Collobert et al. 2011) or by building sequence to sequence predictors based on multiple LSTM layers (Sutskever, Vinyals, and Le 2014). Finally, fully convolutional models exist for such problems as pixel-wise segmentation (Shelhamer, Long, and Darrell 2017). Other deep models rely on deep architectures to define an energy function over candidate labels and then predictions are produced by using backpropagation to iteratively optimize the energy with respect to the labels. Structured Prediction Energy Networks (SPENs) (Belanger and McCallum 2016) take that approach and can capture dependencies that would lead to intractable graphical models. Unfortunately, many of the deep approaches are not probabilistic and are mostly applicable to data of the same global structure and number of output labels as in the training examples.

Sum-Product Networks (SPNs) are unique in being a deep and probabilistic architecture capable of representing high-treewidth models that would result in unmanageable PGMs. Yet, they provide tractable exact inference by exploiting context-specific independence and determinism (Gens and Domingos 2012). In contrast to SPENs and approximate inference over PGMs, marginal and MPE inference in SPNs does not require iterations and can be achieved using a single pass through the network (Poon and Domingos 2011). SPNs are expressive and have been used to solve difficult problems in several domains (Poon and Domingos 2011; Gens and Domingos 2012; Amer and Todorovic 2015), including semantic place classification in robotics (Pronobis and Rao 2017). These benefits resulted in several approaches to structured prediction using SPNs. In (Ratajczak, Tschiatschek, and Pernkopf 2014), Linear-chain Conditional Random Fields (LC-CRFs) extended with local factors modeled using SPNs were used to model sequences. In (Nath and Domingos 2015), Relational-SPNs were proposed that model graph-based relational data based on first-order logic. This method models graphs with potentially varying sizes by summarizing multiple variables with an aggregate statistic. In contrast, we directly model each output variable associated with nodes of the graph, and construct an SPN structure specific to each graph instance.

---

[1] http://coldb.org/graphs

There have been numerous attempts to employ structured prediction to modeling semantic maps with topological spatial relations. In (Mozos et al. 2007), Hidden Markov Models were used to smooth sequences of AdaBoost classifications of place observations into semantic categories. (Friedman, Pasula, and Fox 2007) proposed Voronoi Random Fields (VRFs) which are CRFs constructed according to a Voronoi graph extracted from an occupancy grid map. VRFs utilize pairwise potentials to model dependency between neighboring graph nodes and 4-variable potentials to model junctions. In (Pronobis and Jensfelt 2012), Markov Random Fields were used to model pairwise dependencies between semantic categories of rooms according to a topological map. The categorical variables were connected to Bayesian Networks that reasoned about local environment features, forming a chain graph. This approach relied on a door detector to segment the environment into a topological graph with only one node per room. While these approaches are probabilistic, they rely on approximate inference using Loopy BP leading to problems with convergence (Friedman, Pasula, and Fox 2007). Moreover, in both cases, additional prior knowledge or hard commitments about the semantics of some places were employed in order to obtain a clean and manageable topological graph structure. In contrast, in this work, we rely on a graph built primarily to support navigation and execution of actions by the robot. Such graph provides a better coverage, but results in more noisy structure. Furthermore, we make no hard commitments about the semantics of the places at time of structure creation and defer such inference to the final model. Our experiments show, that under such conditions, graphical models with pairwise or higher-order potentials deteriorate quickly.

## Preliminaries

We begin by giving a brief introduction to SPNs. For details, the reader is referred to (Peharz et al. 2017; Poon and Domingos 2011). Then, we describe the structure of the semantic maps for which we learn GraphSPNs.

### Sum-Product Networks

SPNs are a new promising architecture which combines the advantages of deep learning and probabilistic modeling. One of the primary limitations of traditional probabilistic graphical models is the complexity of their partition function, often requiring complex approximate inference in the presence of non-convex likelihood functions. In contrast, SPNs represent joint or conditional distributions with partition functions that are guaranteed to be tractable and involve a polynomial number of sum and product operations, permitting exact inference. They are capable of learning probabilistic models directly from high-dimensional, noisy data.

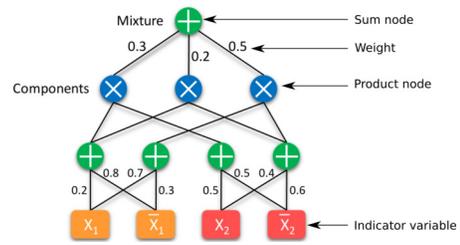

Figure 2: A simple SPN for a naive Bayes mixture model $P(X_1, X_2)$, with 3 components over 2 binary variables.

As shown in Fig. 2, an SPN is a directed acyclic graph composed of weighted sum and product operations. The sums can be seen as mixture models over subsets of variables, with weights representing mixture priors. Products can be viewed as combinations of features. SPNs can be defined for both continuous and discrete variables, with evidence for categorical variables often specified in terms of binary indicators.

Formally, following (Poon and Domingos 2011), we can define an SPN as follows:

**Definition 1** *An SPN over variables* $X_1, \ldots, X_V$ *is a rooted directed acyclic graph whose leaves are the indicators* $(X_1^1, \ldots, X_1^L), \ldots, (X_V^1, \ldots, X_V^L)$ *and whose internal nodes are sums and products. Each edge* $(i, j)$ *emanating from a sum node* $i$ *has a non-negative weight* $w_{ij}$. *The value of a product node is the product of the values of its children. The value of a sum node is* $\sum_{j \in Ch(i)} w_{ij} v_j$, *where* $Ch(i)$ *are the children of* $i$ *and* $v_j$ *is the value of node* $j$. *The value of an SPN* $S[X_1, \ldots, X_V]$ *is the value of its root.*

Not all architectures consisting of sums and products result in a valid probability distribution. While a less constraining condition on validity has been derived in (Poon and Domingos 2011), a simpler condition, which does not limit the power of the model is to guarantee that the SPN is *complete* and *decomposable* (Peharz et al. 2017):

**Definition 2** *An SPN is complete* iff *all children of the same sum node have the same scope.*

**Definition 3** *An SPN is decomposable* iff *no variable appears in more than one child of a product node.*

The scope of a node is defined as the set of variables that have indicators among the descendants of the node.

A valid SPN will compute unnormalized probability of evidence expressed in terms of indicators. However, the weights of each sum can be normalized, in which case the value of the SPN $S[X_1^1, \ldots, X_V^I]$ is equal to the normalized probability $P(X_1, \ldots, X_V)$ of the distribution modeled by the network.

Partial or missing evidence can be expressed by setting the appropriate indicators to 1. Inference is then accomplished by an upwards pass which calculates the probability of the evidence and a downwards pass which obtains gradients for calculating marginals or MPE state of the missing evidence. The latter can be obtained by replacing sum operations with weighted max operations (Peharz et al. 2017).

Parameters of an SPN can be learned generatively (Poon and Domingos 2011) or discriminatively (Gens and Domingos 2012) using Expectation Maximization (EM) or gradient descent. Additionally, several algorithms were proposed for simultaneous learning of network parameters and structure (Hsu, Kalra, and Poupart 2017; Gens and Domingos 2013). In this work, we use a simple structure learning technique (Pronobis and Rao 2017) to build template SPNs. We begin by initializing the SPN with dense structure by recursively generating nodes based on multiple random decompositions of the set of variables into multiple subsets until each subset is a singleton. The resulting structure consists of products combining the subsets in each decomposition and sums mixing different decompositions at each level. Then, we employ hard EM to learn the model parameters, which was shown to work well for generative learning (Pronobis and Rao 2017) and overcomes the diminishing gradient problem. After parameter learning, the generated structure can be pruned by removing edges associated with weights close to zero.

## Semantic Maps

GraphSPNs are applicable to data structured by arbitrary graphs. However, in this work, we apply them specifically to semantic maps employing topological graphs built by a mobile robot exploring a large-scale environment (Pronobis, Riccio, and Rao 2017). The primary purpose of our topological graph is to support the behavior of the robot. As a result, nodes in the graph represent places the robot can visit, and edges represent navigability. Each graph node is associated with a latent variable representing semantic place category and the edges can be seen as spatial relations forming a global semantic map. Local evidence providing information about the semantics of a place might be available and we assume that such evidence is inherently uncertain and noisy. Additional nodes in the graph represent exploration frontiers, i.e. possible places the robot has not yet visited, but can potentially navigate to. We call such nodes *placeholders*, and assume that the robot has not yet obtained any local evidence about their semantics.

The topological graph is assembled incrementally based on dynamically expanding 2D occupancy map. The 2D map is built from laser range data captured by the robot using a Rao-Blackwellized particle filter grid mapping approach (Grisetti, Stachniss, and Burgard 2007). Given the current state of the 2D map, placeholders are added at neighboring, reachable, but unexplored locations and connected to already existing places. Then, once the robot performs an exploration action and navigates to a placeholder, it is converted into a place with local evidence attached.

We formulate the problem of finding new placeholder locations in the 2D occupancy map as continuous sampling from a distribution that models relevance and suitability of the locations. Specifically, the distribution is specified as: $P(E|G, \mathcal{E}) = \frac{1}{Z} \prod_i \phi_S(E_i|G) \phi_N(E_i|\mathcal{E})$, where $E_i \in \{0, 1\}$ represents the existence of a new placeholder at location $i$ in the occupancy map, $G$ is the occupancy grid, and $\mathcal{E}$ is the set of locations of all places that have been added previously. The potential $\phi_S$ ensures that placeholders are located in areas that are safe and preferred for navigation (are within safe distance from obstacles, with the preference towards centrally located places). The potential $\phi_N$ models the neighborhood of a place, and guarantees sufficient coverage of the environment by promoting positions at a certain distance $d_n$ from existing places. Final location of a new placeholder is chosen through MPE inference in $P(E|G, \mathcal{E})$. An edge is then created to represent navigability. It connects the placeholder to an existing place in the graph based on A* search directly over the potential $\phi_S$. An example of such semantic map is shown in Fig. 1.

## GraphSPNs

GraphSPNs build probabilistic models over arbitrary graph-structured data, with local evidence $X_i$ and latent variables $Y_i$ associated with each graph node or edge $i$ and dependencies between the latent variables expressed in terms of the graph structure. A GraphSPN is a template model that is learned from a set of graph-structured data samples, possibly of different global structure. The GraphSPN is then instantiated for a specific data structure $D = (G, X, Y)$, where $G = (V, E)$ is a graph with vertices $V$ and edges $E$, and $X = \{X_i : i \in V \cup E\}, Y = \{Y_i : i \in V \cup E\}$. In our specific implementation of GraphSPNs for semantic maps, $G$ is assumed to be the current topological graph, $X_i$ is the local evidence about the semantics of each place, and $Y_i$ captures the latent semantic place category.

In order to learn a GraphSPN, we begin by specifying a set $\mathcal{G}$ of *sub-graph templates*, which are used to decompose any data structure $D$. We define such decomposition as follows:

**Definition 4** *A decomposition of a data structure $D = (G, X, Y)$ using sub-graph templates $\mathcal{G}$ is a set of components $D_k = (G_k, X_k, Y_k)$, with $G_k = (V_k, E_k)$, such that $G_k$ is isomorphic with any $\mathcal{G} \in \mathcal{G}$, $\bigcup_k G_k = G$, $\forall_{k,l} G_k \cap G_l = \emptyset$, and the variables $X_k$ and $Y_k$ correspond to vertices and edges of $G_k$: $X_k = \{X_i : i \in V_k \cup E_k\}$, $Y_k = \{Y_i : i \in V_k \cup E_k\}$.*

A GraphSPN is a template model consisting of *template SPNs*, one for each sub-graph template, with structure and parameters learned from training samples. Specifically:

**Definition 5** *GraphSPN $\mathcal{S}^\mathcal{G}$ is a set of template SPNs $\mathcal{S}^\mathcal{G}[\mathcal{X}, \mathcal{Y}] \in \mathcal{S}^\mathcal{G}$ corresponding to sub-graph templates $\mathcal{G} \in \mathcal{G}$, with $\mathcal{X}$ and $\mathcal{Y}$ representing local evidence and latent variables associated with vertices and edges of $\mathcal{G} = (\mathcal{V}, \mathcal{E})$: $\mathcal{X} = \{\mathcal{X}_i : i \in \mathcal{V} \cup \mathcal{E}\}, \mathcal{Y} = \{\mathcal{Y}_i : i \in \mathcal{V} \cup \mathcal{E}\}$.*

A GraphSPN $\mathcal{S}^\mathcal{G}$ is *instantiated* for a specific data structure $D = (G, X, Y)$ to obtain an SPN $S_D^\mathcal{G}[X, Y]$ modeling the distribution $P_D^\mathcal{G}(X, Y)$. The process is formalized in Alg. 1 and the resulting SPN is illustrated in Fig. 3. First, the data structure is decomposed multiple times, each time differently, into multiple *components* using the sub-graph templates $\mathcal{G}$. The process of generating the decompositions in our implementation is shown in Alg. 2. We generate different decompositions randomly, each time prioritizing sub-graph templates of higher complexity (larger size). For each decomposition and each component, the corresponding template SPN is instantiated for random variables associated

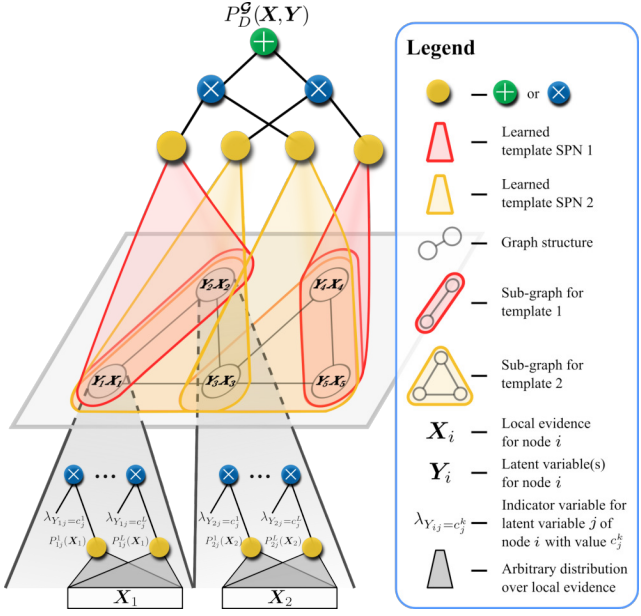

Figure 3: An instantiation of GraphSPN for data structured by a 5-node graph (gray) with variables $X_i$ and $Y_i$ associated with graph nodes. Two different decompositions of the graph structure with two sub-graph templates are shown.

---

**Algorithm 1:** InstantiateGraphSPN($D, \mathcal{G}, S^{\mathcal{G}}$)

1 $\mathring{R} \leftarrow$ Sum node without children;
2 **foreach** $d \in \{1, 2, \ldots, N_D\}$ **do**
3  $\quad \mathring{P}_d \leftarrow$ Product node without children;
4  $\quad \boldsymbol{D} \leftarrow$ DecomposeDataStructure($D, \mathcal{G}$);
5  $\quad$ **foreach** $D_k = (G_k, \boldsymbol{X}_k, \boldsymbol{Y}_k) \in \boldsymbol{D}$ **do**
6   $\quad\quad \mathcal{G} \leftarrow$ Sub-graph template isomorphic with $G_k$;
7   $\quad\quad \mathcal{S}^{\mathcal{G}}[\mathcal{X}, \mathcal{Y}] \leftarrow$ Template SPN for $\mathcal{G}$ from $\boldsymbol{S}^{\mathcal{G}}$;
8   $\quad\quad S^{\mathcal{G}}[\boldsymbol{X}_k, \boldsymbol{Y}_k] \leftarrow$ Instantiate $\mathcal{S}^{\mathcal{G}}$ for $\boldsymbol{X}_k, \boldsymbol{Y}_k$;
9   $\quad\quad$ Add $S^{\mathcal{G}}[\boldsymbol{X}_k, \boldsymbol{Y}_k]$ as a child of $\mathring{P}_d$;
10 $\quad$ **end**
11 $\quad$ Add $\mathring{P}_d$ as a child of $\mathring{R}$;
12 **end**
13 **return** SPN $S_D^{\mathcal{G}}[\boldsymbol{X}, \boldsymbol{Y}]$ rooted in $\mathring{R}$;

---

**Algorithm 2:** DecomposeDataStructure($D, \mathcal{G}$)

1 $\boldsymbol{D} \leftarrow \emptyset$;
2 $D' = (G', \boldsymbol{X}', \boldsymbol{Y}') \leftarrow$ Copy of $D$;
3 **foreach** $\mathcal{G} \in \boldsymbol{\mathcal{G}}$ *in decreasing level of complexity* **do**
4  $\quad$ **while** $\exists_{G_k \in G'} G_k$ *isomorphic with* $\mathcal{G}$ **do**
5   $\quad\quad G_k = (\boldsymbol{V}_k, \boldsymbol{E}_k) \leftarrow$ Random sub-graph of $G'$
6   $\quad\quad\quad\quad\quad\quad\quad$ isomorphic with $\mathcal{G}$;
7   $\quad\quad \boldsymbol{X}_k \leftarrow \{\boldsymbol{X}_i : i \in \boldsymbol{V}_k \cup \boldsymbol{E}_k\}$;
8   $\quad\quad \boldsymbol{Y}_k \leftarrow \{\boldsymbol{Y}_i : i \in \boldsymbol{V}_k \cup \boldsymbol{E}_k\}$;
9   $\quad\quad D_k \leftarrow (G_k, \boldsymbol{X}_k, \boldsymbol{Y}_k)$;
10  $\quad\quad \boldsymbol{D} \leftarrow \boldsymbol{D} \cup D_k$;
11  $\quad\quad G' \leftarrow G' \setminus G_k$;
12 $\quad$ **end**
13 **end**
14 **return** *Decomposition* $\boldsymbol{D}$;

---

with the component. This results in multiple sub-SPNs sharing weights and structure. The instantiated sub-SPNs for a single graph decomposition are combined with a product node, and the product nodes for all decompositions become children of the root sum node. This forms a complete distribution, which can be seen as a mixture model over the different decompositions. Since components of each decomposition do not overlap and cover the complete graph structure, the instantiation of GraphSPN is guaranteed to be decomposable and complete, therefore valid. Additionally, due to the way the SPN is constructed, it is easy to perform an incremental update reflecting a possible change in the data structure. The resulting joint distribution $P_D^{\mathcal{G}}(\boldsymbol{X}, \boldsymbol{Y})$ can be used to perform various types of inferences over the evidence $\boldsymbol{X}$ and latent variables $\boldsymbol{Y}$.

The structures and parameters of *template SPNs* can be learned directly from training data samples decomposed into components. Each template SPN is then trained separately, using only its corresponding components. In order to incorporate the latent variables $\boldsymbol{Y_i} = [Y_{ij}]$, we include a fixed intermediate layer of product nodes into the structure of template SPNs. As shown in Fig. 3, each such product node combines an arbitrary distribution $P_{ij}^k(\boldsymbol{X}_i)$ with an indicator $\lambda_{Y_{ij}=c_j^k}$ for a specific value $c_j^k$ of $Y_{ij}$. The structure of the template SPN built on top of the product nodes can be learned from data and the distributions $P_{ij}^k(\boldsymbol{X}_i)$ can be arbitrary, e.g. realized with an SPN with data-driven structure.

In our experiments, we assumed only one latent variable (semantic place category) $Y_i$ per graph node $i$, with $Val(Y_i) = \{c^1, \ldots, c^L\}$, and we defined $P_i^k(\boldsymbol{X}_i)$ for a single hypothetical binary local observation $x_i$, which we as-

sumed to be observed:

$$P_i^k(X_i) = \begin{cases} \alpha_i^k & X_i = x_i \\ 1 - \alpha_i^k & X_i = \bar{x}_i \end{cases} \quad (1)$$

Such simplification allows us to thoroughly evaluate GraphSPNs for the problem of learning semantic maps by directly simulating hypothetical local evidence about the semantic category of varying uncertainty and under various noise conditions. Furthermore, it allows us to compare GraphSPNs with Markov Random Fields using the same $\alpha_i^k$ as the value of local potentials, i.e. $\phi_i(Y_i = c^k) = \alpha_i^k$. The proposed approach naturally extends to the case where a more complex distribution is used to model semantic place categories based on robot observations, such as the SPN-based approach presented in (Pronobis and Rao 2017). Note, that we still learn the structure and parameters of the *template SPNs* built on top of the distributions $P_i^k(X_i)$.

## Experimental Procedure

### Dataset

The semantic maps used in our experiments were obtained by deploying our semantic mapping framework on sequences of laser range data and odometry captured by a mobile robot exploring multiple large-scale environments (Pronobis and Caputo 2009). The dataset contains 99 sequences, and as a result 99 topological graphs, captured on 11 floors of 3 buildings in different cities. We identified 10 semantic place classes that are common for all buildings (e.g. a corridor, a doorway, a 1-person office, see Fig. 8 for a complete list) and annotated each topological graph node with its groundtruth class.

| NL | $D_{groundtruth}$ | $D_{incorrect}$ |
|----|-------------------|-----------------|
| 1 | 0.991 (+/-0.001) | 0.0 |
| 2 | 0.913 (+/-0.015) | 0.085 (+/- 0.056) |
| 3 | 0.720 (+/-0.040) | 0.090 (+/- 0.061) |
| 4 | 0.434 (+/-0.054) | 0.092 (+/-0.062) |
| 5 | 0.316 (+/-0.030) | 0.154 (+/-0.055) |
| 6 | 0.154 (+/-0.021) | 0.217 (+/-0.074) |

Table 1: Noise levels used in our experiments.

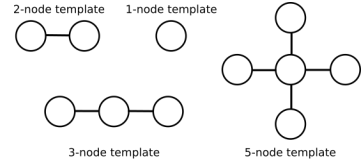

Figure 4: Sub-graph templates used in our experiments.

Our goal in this work is to evaluate the ability of GraphSPNs to disambiguate semantic place classes despite noisy and uncertain local evidence by exploiting spatial relations captured in a noisy topological graph. Probabilistic place classification algorithms, such as the SPN-based approach in (Pronobis and Rao 2017) associate decisions based on local observations with probability estimates. However, the certainty of a decision can be low or the decision can be incorrect. In order to measure how sensitive the evaluated approaches are to such noise, we simulate local evidence attached to topological graph nodes by adding increasing noise to groundtruth information.

To this end, for each node $i$ in each topological graph, we generated a local evidence distribution with values $P(Y_i = c^k, X_i = x_i) = \alpha_i^k$. For each graph, we first randomly selected 20% of all nodes for which the most likely local result should be incorrect. For those, we selected a random incorrect class to be associated with the highest probability value. Then, we randomized the value $D_{incorrect}$, which is a difference between the highest probability and the probability of the groundtruth class, from a uniform distribution in a range depending on the noise level. For the remaining 80% of nodes, we ensured that the groundtruth class is associated with the highest probability. However, we simulated uncertainty by randomizing the value $D_{groundtruth}$, which is a difference between the probability of the groundtruth class and the second highest probability. With these constraints, we used random values for the remaining likelihoods and made sure that each distribution is normalized. Intuitively, lower $D_{groundtruth}$ indicates higher uncertainty and higher $D_{incorrect}$ indicates stronger noise. The statistics of the values of $D_{incorrect}$ and $D_{groundtruth}$ for the final evidence at different noise levels are shown in Tab. 1.

### Learning GraphSPNs

We learned GraphSPNs from a simple set of *sub-graph templates* shown in Fig. 4, matching from 1 to 5 nodes and simple edge configurations. We assumed that each node is associated with a single latent variable $Y_i$ representing the semantic place class. For each sub-graph template with at least 2 nodes, we learned a *template SPN* of specific structure and parameters from components of annotated training semantic maps corresponding to the sub-graph template ($Y_i$ was set to the groundtruth). Each training graph was decomposed in 10 different ways to obtain components. For a single-node template we simply assumed a uniform SPN. During testing, we built the *instance GraphSPN* based on 5 different graph decompositions (see Fig. 8). In our experiments, we always learned GraphSPN from all graphs from two buildings in the dataset and tested on graphs with different evidence noise levels from the remaining building. GraphSPNs were implemented using LibSPN (Pronobis, Ranganath, and Rao 2017).

### Constructing Markov Random Fields

We compared GraphSPNs to a traditional approach based on MRFs structured according to the represented graph. The MRF was constructed from two types of potentials: potential $\phi_i(Y_i = c^k) = \alpha_i^k$ used to provide local evidence, and potentials modeling latent variable dependencies. For the latter, we tried two models: using pairwise potentials for each pair of variables associated with connected nodes or defined over three variables for three connected nodes in any configuration. In each case, the potentials were obtained by generating co-occurrence statistics of variable values in the training graphs used for learning GraphSPNs. Inference in the MRF was performed using Loopy BP implemented in the libDAI library (Mooij 2010).

### Experimental Results

We performed several experiments comparing the learned GraphSPN model to the MRF models with pairwise potentials (marked as MRF-2) and three-variable potentials (marked as MRF-3). First, we tasked all models with disambiguating noisy local evidence about semantic place class for places visited by the robot. For each topological graph in the test set, we performed marginal inference[2], based on which we selected the final classification result as $\mathrm{argmax}_k P_D^{\mathcal{G}}(Y_i = c^k | \boldsymbol{X} = \boldsymbol{x})$.

The percentage of correctly classified nodes in the graph averaged over all test graphs is shown in Fig. 5 and a visualization of results for a single graph together with the decompositions used to build the *instance GraphSPN* is shown in Fig. 8. We evaluated all assignments of the three buildings into training and test sets as well as different noise levels listed in Tab. 1. Each test set consisted of 5 topological maps, each with 3 random sets of noisy local evidence resulting in 15 different test graphs. Since the local evidence for 20% of nodes in each graph indicates an incorrect class as the most likely one, only accuracy greater than 80% demonstrates that the model was able to recover from the noise using learned spatial relations. Lower accuracy suggests that the incorrect evidence was too strong or that the correct evidence was too uncertain to influence the semantic class of a place.

---
[2]We experimented with MPE inference over all latent variables achieving inferior results with all models.

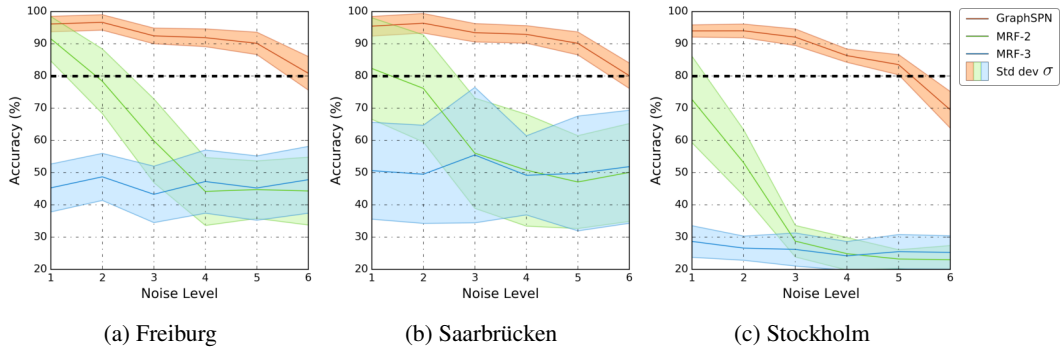

(a) Freiburg  (b) Saarbrücken  (c) Stockholm

Figure 5: Semantic place classification accuracy for all models and test buildings, and at different noise levels.

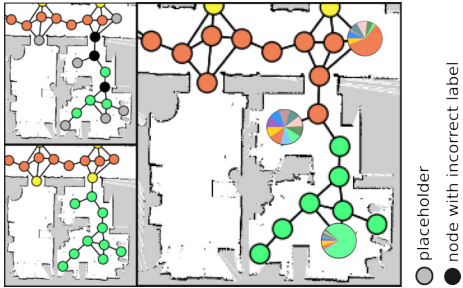

Figure 6: Visualization of results for the experiment involving placeholders without local evidence, at noise level 2. Left, bottom: the semantic map with groundtruth semantic place classes (including for placeholders). Left, top: the 20% of nodes for which the most likely evidence indicates an incorrect class (black) and placeholders with no evidence (gray). Right: inferred marginal distributions over semantic classes of placeholders (pie charts).

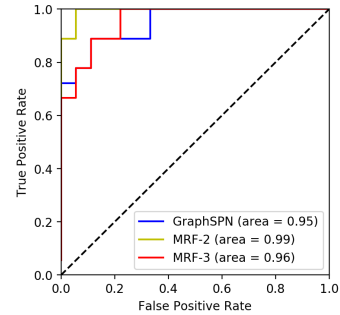

Figure 7: ROC curve for the novelty detection task with certain semantic label assignments.

Analyzing the performance reported in Fig. 5, we see that pairwise MRF performs well when there is little noise in the evidence, however it deteriorates quickly with increasing noise levels. At the same time, GraphSPN generates robust results (accuracy greater than 80%) despite substantial noise and uncertainty. With substantial noise, approximate Loopy BP inference for MRF converges to a solution consisting primarily of the dominant class (the corridor). At the same time, we see that using higher-order potentials with MRF actually hurts performance.

In the second experiment, we tasked the models with inferring marginal distributions over the semantic classes of places not yet visited by the robot (placeholders) for which local evidence is unavailable. We used the same setup as in the previous experiment, with local evidence for explored places including noise and uncertainty. Examples of such marginal distributions are shown in Fig. 6, while the classification accuracy when considering the most likely class is reported in Tab. 2. Again, GraphSPN significantly outperformed the MRF for this inference task. If we analyze the marginal distributions for three representative placeholders shown in Fig. 6, we see that GraphSPN is confident about the correct class for the two placeholders for which nearby nodes provide correct (albeit uncertain) evidence. For the placeholder connected to nodes for which evidence indicates an incorrect class the marginal distribution is almost uniform. This indicates the ability of GraphSPN to generate useful confidence signal in presence of noisy inputs.

Our final experiment was designed to evaluate the quality of likelihood produced by GraphSPN instantiated over a graph of arbitrary size. In this experiment, we tasked the models with detecting novel spatial environment configurations that are incongruent with the evidence available during training. To this end, we produced novel semantic maps by swapping groundtruth semantic class labels between corridors and doorways as well as cooridors and 1-person offices. We contrasted these graphs with the original groundtruth as well as swaps that are consistent with the training data (swapping 1-person offices with 2-person offices). To detect novel configurations, we measured the likelihood of the complete graph with certain label assignments (without any noise) and thresholded the obtained likelihood to decide whether a graph is likely to be generated from the distribution. Since the likelihood depends on the size of the graph, we normalized it by the number of variables in the test graph before performing thresholding. We used the models learned in the previous experiments and produced false and true positive rates for various threshold values over all trained models and test sets.

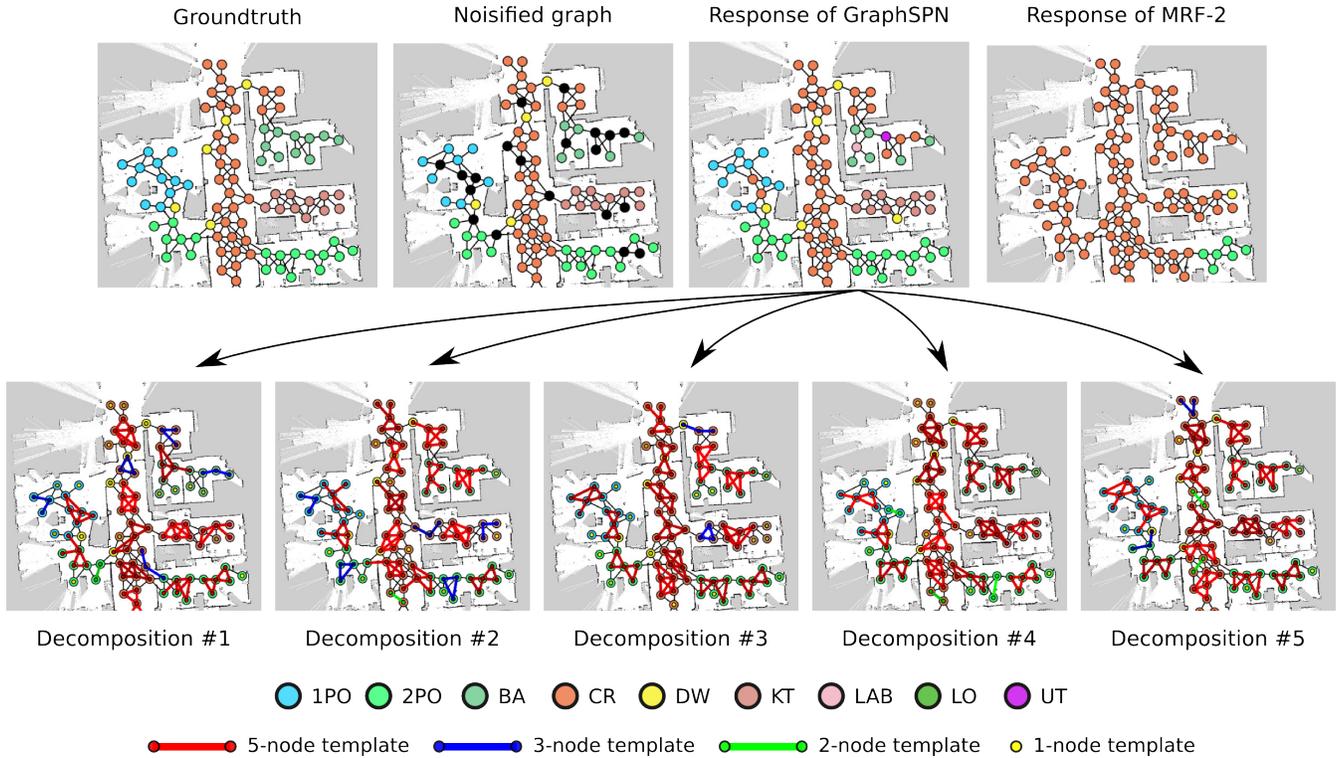

Figure 8: Visualization of the results for a graph from Freiburg at noise level 4. The top row shows: the semantic map with groundtruth semantic place classes; the 20% of nodes for which the most likely evidence indicates an incorrect class (black nodes); the semantic classes inferred by the GraphSPN and the MRF-2. The place classes are: 1-person office (1PO), 2-person office (2PO), bathroom (BA), corridor (CR), doorway (DW), kitchen (KT), laboratory (LAB), large office (LO), meeting room (MR), utility room (UT). The bottom row illustrates the 5 decompositions used when assembling the *instance GraphSPN* (different colors indicate different sub-graph templates applied).

| | GraphSPN | | |
|---|---|---|---|
| NL | Freiburg | Saarbrücken | Stockholm |
| 2 | 67.58%(+/-10.42) | 78.15%(+/-9.95) | 67.57%(+/-11.11) |
| 5 | 40.59%(+/-12.22) | 55.18%(+/-19.67) | 37.56%(+/-10.44) |
| | MRF-2 | | |
| NL | Freiburg | Saarbrücken | Stockholm |
| 2 | 28.32%(+/-7.53) | 39.85%(+/-19.42) | 12.44%(+/-3.46) |
| 5 | 24.23%(+/-11.40) | 30.58%(+/-5.57) | 10.04%(+/-2.59) |
| | MRF-3 | | |
| NL | Freiburg | Saarbrücken | Stockholm |
| 2 | 28.71%(+/-5.43) | 31.94%(+/-5.26) | 10.11%(+/-0.51) |
| 5 | 18.02%(+/-7.49) | 28.86%(+/-6.16) | 8.96%(+/-1.19) |

Table 2: Accuracy of semantic class inference for placeholders without local evidence, for all models and test buildings, and at two representative noise levels.

The results for such detection task are shown as ROC curves in Fig. 7. All three models perform well on this task (as measured by the area under the curve). We see that GraphSPN performs comparably to the MRF models demonstrating its ability to produce useful likelihood values. At the same time, MRFs perform slightly better on this task. This is a result of lack of noise and uncertainty in the provided evidence. It shows that MRFs can capture relevant spatial relations and suggests that their performance deteriorates due to the limitations of approximate inference.

## Conclusions

We presented GraphSPNs, a probabilistic deep model for graph-structured data that learns a template distribution allowing for making inferences about graphs of different global structure. While existing works applied SPNs to data organized as fixed-size grids or sequences, this paper presents a novel attempt at deploying SPNs on arbitrary graphs of varying size. Based on GraphSPNs, we proposed a method for learning topological spatial relations in semantic maps constructed by a mobile robot. Our method is robust to noise and uncertainty inherent in real-world problems where information about the environment is captured with robot sensors. Our framework is universal and compat-ible with any distributions defined over local evidence. However, it is particularly well suited for integration with other SPN-based models. In the future, by combining a GraphSPN learned over semantic maps with the generative place model proposed in (Pronobis and Rao 2017), we intend to achieve a unified, deep, and hierarchical representation of spatial knowledge spanning from local sensory observations to global conceptual descriptions.